# Instance Segmentation of Reinforced Concrete Bridges with Synthetic Point Clouds

Asad Ur Rahman[a], Vedhus Hoskere[a*]

[a] *Department of Civil and Environmental Engineering, University of Houston, 4226 MLK Blvd, Houston, TX 77204, United States*

* Corresponding author at: Department of Civil and Environmental Engineering, University of Houston, *4226 MLK Blvd, Houston, TX 77204, United States*

E-mail addresses: aurahman@cougarnet.uh.edu (A. Rahman), vhoskere@central.uh.edu (V. Hoskere).

**Abstract**

Bridge inspections are essential to ensure the safety and structural integrity of these vital transportation links. The National Bridge Inspection Standards (NBIS) mandate detailed element-level bridge inspections every two years, requiring states to assess and report the condition of individual bridge components. Traditionally, inspectors manually assign condition ratings by rating structural components based on visual damage, but this process labor-intensive, time-consuming, and has limited efficacy. Improving the element-level bridge inspection process with automated data collection and processing can facilitate more comprehensive condition documentation and enable informed decision making to improve overall bridge management. The accurate identification of individual bridge elements is an important processing task towards such automation. While researchers have extensively studied the semantic segmentation of bridge point clouds, there has been limited research on the instance segmentation of bridge elements. Achieving accurate element-level instance segmentation requires a large amount of annotated bridge point clouds to train deep learning-based instance segmentation models but such datasets are difficult to acquire and label. To address the lack of such datasets, we propose a novel approach for the generation of synthetic data of structural components. We arrive at our data generation approach by evaluating three distinct sampling methods. Further, we propose a framework for deep learning-based instance segmentation of real RC bridge point clouds using models trained on data produced from our synthetic data generation approaches. We leverage the Mask3D transformer-based model, exploring various training strategies, including hyperparameter tuning and a novel sparsity-based point cloud occlusion pre-processing technique. Finally, we demonstrate the capability of a model trained with our framework to achieve an mAP of 97.8% and 63.8% on real LiDAR and photogrammetry RC bridge point clouds that we acquire for the purposes of this study respectively. This research underscores the value of scientific and systematic synthetic data generation and presents a framework that can be utilized at scale towards identification of individual bridge elements for automating the element-level bridge inspections.

1. Introduction

Bridges are essential transportation infrastructure that connect communities and enable economic growth, making their continued operation, and thus their continued inspection, a matter of paramount importance [1]. Bridges are constantly exposed to varying vehicular, human, and environmental loading, all of which can lead to deterioration over time. Bridge performance can also deteriorate due to design or construction flaws yielding catastrophic results [2]. To facilitate continued operation, the National Bridge Inspection Standards (NBIS) mandate that all bridges must undergo condition assessment every two years [3]. However, NBIS inspections are typically conducted manually and can be inefficient, laborious, expensive, and dangerous and involve assigning condition ratings to each bridge element.

Researchers have sought to digitize the bridge inspection process through the creation of digital twins that can serve as a reliable surrogate of a physical bridge and be used for decision making. In essence, digital twins are digital replicas of the infrastructure asset [4–8] that contain information about the structure and its condition. A first step in developing a DT is the production of a reality model which is 3D representation of the asset produced using photogrammetry or LiDAR using collected raw data. The reality model is then processed to obtain relevant semantic representations and information relating to structural components and damage. Much research has been devoted to the identification of bridge damage using images [9–13] and so we focus here on the less studied problem of identifying bridge components using point clouds of reality models.

To extract information about bridge components from point clouds, 3D semantic segmentation approaches have been researched. Studies related to semantic segmentation of bridge point clouds can be divided into two categories, (i) heuristic methods [14–18], and (ii) learning based methods [19–25]. Xia et al. [20] proposed a multi-scale local descriptor and machine learning-based approach for semantic segmentation of bridge components. Their method outperformed PointNet on the real-world dataset by Lu et al. [14]. Yang et al. [21] adopted the weighted superpoint graph (WSPG) approach for the semantic segmentation of bridge components. The study claimed superior performance of the WSPG model compared to PointNet, DGNN, and superpoint graph

(SPG). Lin et al. [23] proposed a framework for the semantic segmentation of road infrastructure bridge components using a mobile LIDAR mapping system. This methodology, evaluated on 27 interstate highway bridges, employed the spearpoint graph (SPG) approach for semantic segmentation, categorizing bridge and road infrastructure into 9 distinct classes. J. S. Lee et al. [24] utilized a hierarchical Dynamic Graph-based Convolutional Neural Network (DGCNN) for semantic segmentation of railway bridges, reporting improved performance, particularly in the vicinity of tall electric poles near the bridge. Yang et al. [25], introduced a framework for semantic segmentation of bridge point clouds. The authors proposed two synthetic data augmentation techniques and employed a graph structured deep metric learning approach to enhance the weighted spearpoint graph (WSPG) model. Most recent study by Shi et al. [26] developed a method for generating large-scale synthetic 3D point cloud datasets. The approach includes generating random bridge types, simulating camera trajectories, using Structure from Motion (SfM) for 3D reconstruction, and training 3D semantic segmentation models for bridge components segmentation.

The semantic segmentation of bridge components is useful for the identification of bridge component class of every point in the point cloud. However, to produce digital twins and the automate bridge element rating, additional information is required about the location individual bridge elements, also termed as instances of the bridge components. Semantically segmented point clouds thus still require further processing to identify individual instances of any component type. We illustrate a workflow for digital twin generation using instance segmentation in Figure 1. While reviewing existing research, we discovered only two studies about instance segmentation of bridge point clouds [18,19]. These studies specifically focus on truss bridges. Lamas D et al. [19] uses deep learning, while Lamas D et al. [18] employs a heuristic method involving Principal component analysis (PCA). There has been no research on instance segmentation of reinforced concrete bridges which constitute the bulk of the bridge type in the world.

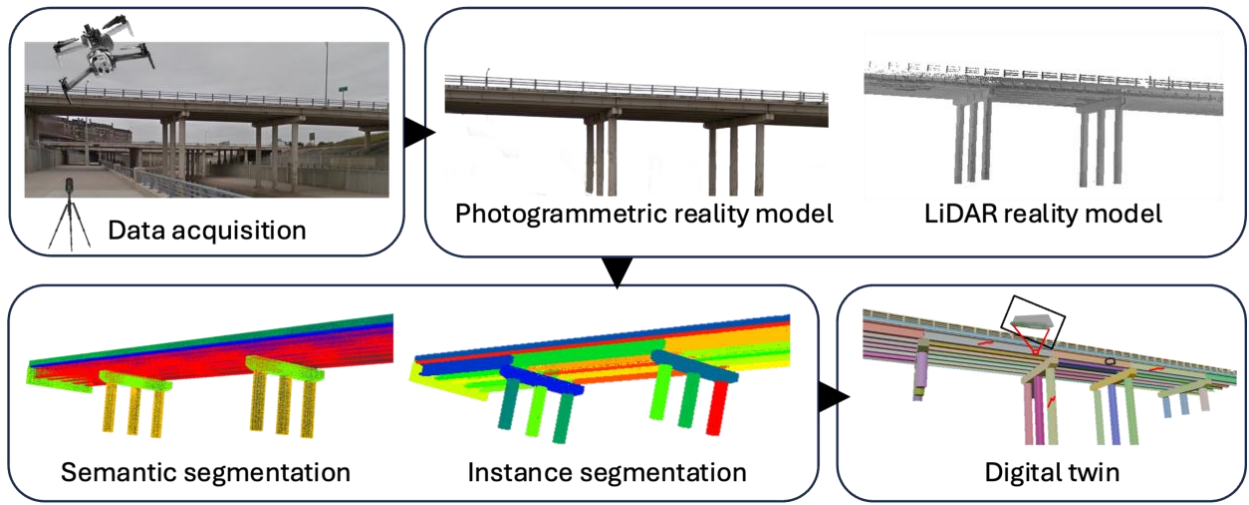

Figure 1: Workflow for digital twin creation starting with data acquisition through photogrammetry and LiDAR scanning. The process includes the development of a reality model, semantic and instance segmentation of structural components, and the identification of damages such as cracks and deflections

There has recently been an explosion of research on developing instance segmentation methods for other applications that could potentially be adopted to improve instance segmentation of bridge points clouds. Methods include top-down proposal-based methods [27–31], bottom-up grouping-based methods [32–36], and voting-based approaches [37–41]. The SoftGroup [38] has two-stage pipeline, benefiting from both proposal-based and grouping-based approaches. In the bottom-up stage, high-quality object proposals are generated through soft semantic score grouping, and in the top-down stage, each proposal is processed to refine positive samples and suppress negative ones. Recently, state of the art transformer-based approaches that outperformed the previous studies include SPFormer [42], Mask3D [43], and OneFormer3D [44]. SPFormer [42] is two-stage approach, grouping potential point features from Sparse 3D U-Net into superpoints. Instances are predicted through a novel query-based transformer decoder. Mask3D [43] uses sparse 3D U-Net based backbone for feature extraction, followed by transformer decoder with separate queries for each instance. The instance queries are learned by iteratively attending to the feature maps at multiple levels. OneFormer3D [44] unifies the semantic, instance and panoptic segmentation of 3D point cloud in a single model. Training models for instance structural components requires a huge amount of labeled point cloud data. Acquiring labeled point cloud for training purpose is very expensive [45]. Therefore, researchers have directed their efforts towards the generation of synthetic point clouds [19,46–49] for various tasks.

This study proposes a novel framework for the instance segmentation of reinforced concrete bridge point clouds using synthetic point clouds and state-of-the art transformer model. Specifically, the three main contributions of the framework are as follows. First, we propose a novel approach for generating synthetic point clouds dataset of reinforced concrete bridges with the aim of facilitating instance segmentation of structural components from point clouds of reality models. Second, we propose a novel sparsity-based occlusion algorithm for data augmentation of bridge point clouds that improves generalizability of models trained on synthetic data. Finally, we carefully evaluate the applicability of Mask3D model for the task of instance segmentation of bridge components by conducting experiments for optimal hyperparameter selection. We also studied the effect of various synthetic dataset types and the effect of occlusion on instance segmentation of structural components of bridges.

The structure of this work is organized as follows. Section 3 proposed methodology, section 4 discusses about the training and evaluation, section 5 provides a comprehensive analysis of the results from the experimentation. Finally, Section 6 and section 7 gives the conclusion and limitations.

## 2. Methodology

This section outlines the methodology crafted to execute the instance segmentation of RC bridges. Our proposed methodology is illustrated in Figure 2, consisting of five steps namely (i) synthetic RC bridge point clouds generation, (ii) data pre-processing, (iii) data augmentation strategies, and finally (iv) instance segmentation model, and (v) field data collection conducted to validate the proposed approach (section 3.4).

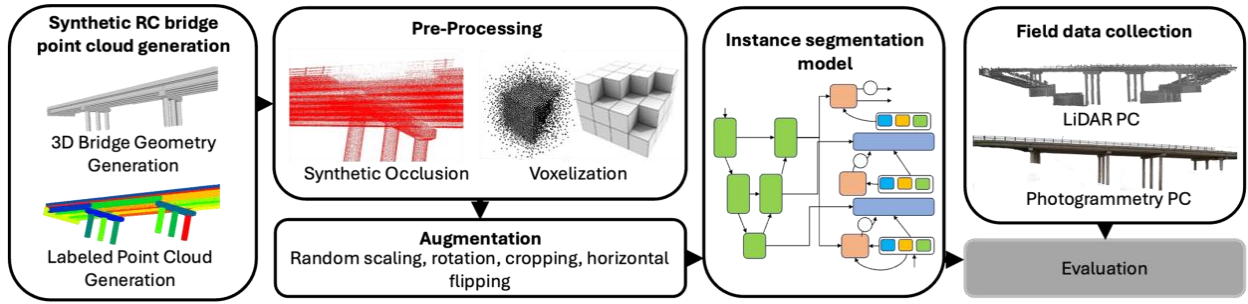

Figure 2. Diagram illustrates the training and testing pipeline for instance segmentation on bridge point clouds. Synthetic data generation followed by pre-processing, data augmentation, and training the deep learning-based instance segmentation model. The real-world data (both LiDAR and photogrammetry point clouds) are employed for testing the model's performance.

## 2.1. Synthetic RC Bridge Point Clouds Generation

The lack of labeled RC bridge point cloud dataset necessitates the development of synthetic bridge dataset for training the instance segmentation model. The existing dataset by Lu et al. [14], which only includes point clouds for 10 bridges without labels, is insufficient for deep learning models. A synthetic labeled data featuring diverse bridge designs, is critical for developing models capable of accurately segmenting bridge point clouds.

### 2.1.1. 3D Bridge Geometry Generation

The initial steps common to creating all three synthetic datasets involve 3D modeling and processing. Using Open bridge designer (OBD), 60 bridges were modeled with randomly defined cross-sections (Figure 3) for their structural components, employing RC bridge modeling templates for rapid development. These bridge models, focusing on above-ground structural components like slabs, barriers, girders, pier caps, and piers, for instance segmentation purposes, were exported in .obj format. Some of the bridge models' examples are illustrated in the Figure 4. Notably, a few synthetic bridges that were roughly similar to the real test bridge (though not identical in member and cross-sectional dimensions) were included to ensure comprehensive coverage of the real bridge type in the training data distribution. After exporting the bridge model from Open Bridge Modeler (OBM), it undergoes further processing in Blender, where unnecessary or invisible parts, such as wing walls, footings, and piles, are removed to focus on relevant segment classes under study. The model is then centered to the origin, aligned, and its layers are split into separate components, preparing it for the specific requirements of simulation within the IsaacSim

Omniverse environment. These processing steps are crucial for ensuring the model's compatibility in subsequent simulation environment.

*2.1.2. Point cloud generation approaches*

We generate synthetic RC bridge point clouds using two different approaches as shown in the Figure 5: (i) Mesh Sampled Point cloud (MSP), (ii) Simulated LiDAR Point Cloud (RSLP).

2.1.2.1. Mesh Sampled Point Cloud (MSP)

Mesh Sampled Point Cloud (MSP) dataset was generated by sampling points on the 3D meshes of each bridge component using CloudCompare. During preprocessing, the mesh was segmented into distinct structural components based on semantic and instance classifications utilizing the Blender Python library. Point clouds for each component were created through density-based sampling on the meshes. Subsequently, these point clouds were labeled according to their respective components and merged into a comprehensive single bridge point cloud. This process was fully automated across all bridges. Unlike the real world LiDAR point cloud, which can only capture external surfaces visible to scanner and leave occluded areas like the interior of hollow sections and contact points undetected, mesh sampled point cloud from mesh models using cloud compare include detailed internal geometries (Figure 6).

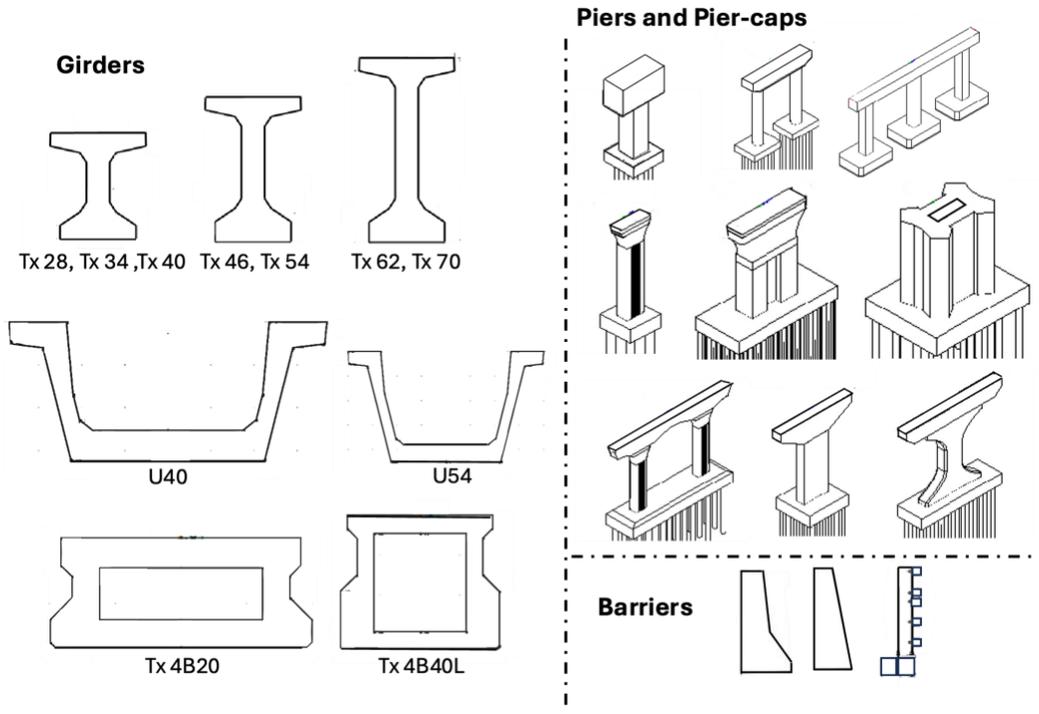

Figure 3. shows various cross-sections of bridge elements used to create bridge models.

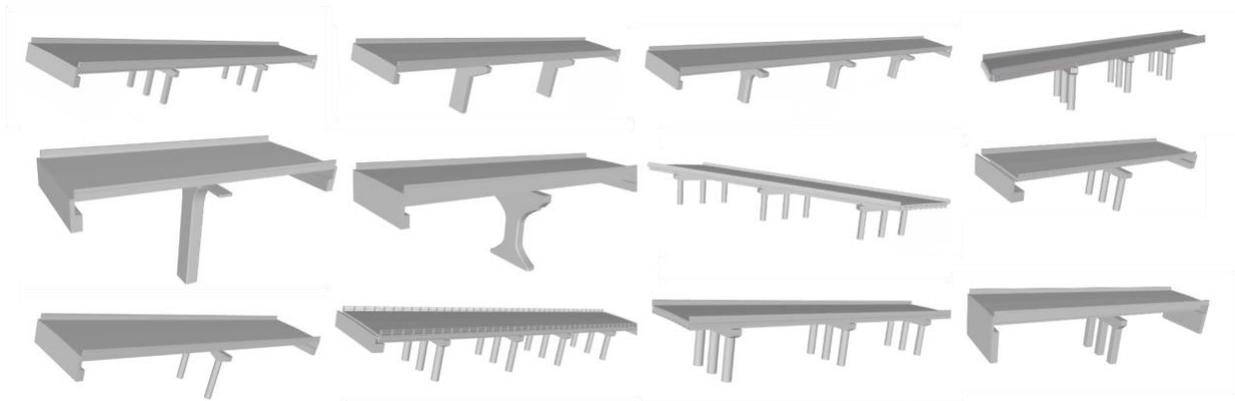

Figure 4. Displays various structural configurations of synthetic bridge geometries modeled in OpenBridge designer.

2.1.2.2.   Simulated LiDAR Point cloud (SLP)

To address the differences between mesh sampled point clouds and those obtained from real-world LiDAR, a simulated environment was established using IsaacSim Omniverse, where LiDAR sensors were strategically placed around the bridge to enhance point cloud coverage (Figure 5). These sensors were configured with a minimum and maximum range of 0m to 600m, and both the horizontal and vertical fields of view were set at 360º and 180º, respectively, with resolutions of 0.4º. Using the LiDAR Python API, semantic labels were applied to the point clouds, correlating with object layers for structural components such as slabs, barriers, girders, pier caps, and piers. Each structural component of the bridge was also given a unique instance label. Due to the absence of texture or color information in the bridge models, all points were assigned RGB values of 255 (white), representing a uniform coloration. The comprehensive point cloud data was ultimately saved in a .txt file format, containing fields for coordinates (X, Y, Z), color values (R, G, B), and both semantic (sem) and instance (inst) labels. The synthetic bridges were simulated in the environment with different sensor configurations, resulting in two different datasets: the Realistic Simulated LiDAR Point Cloud (RSLP) and the Perfect Simulated LiDAR Point Cloud (PSLP).

**Realistic Simulated LiDAR Point Cloud (RSLP)**

The LiDAR sensors were initially placed in a realistic and practical manner, considering operator accessibility in the field (Figure 5). Consequently, 12 LiDAR sensors—six above the bridge deck and six at ground level—were strategically placed to enhance point cloud coverage. It was observed that the LiDAR-generated point cloud omitted some occluded areas, such as those between closely placed girders, highlighting the inherent limitations of LiDAR simulations in capturing complex structural details (Figure 6).

**Perfect Simulated LiDAR Point Cloud (PSLP)**

To address the limitation of missing critical parts with conventional LiDAR setups, we designed a more comprehensive sensor arrangement that, while impractical for real-world applications, ensures no part of the bridge, including occluded areas, is overlooked. This time the LiDAR sensors were deployed in a dense 3D grid with four levels—two above and two below the bridge— tailored to the bridge's size. This configuration varied from 4 to 6 rows and 8 to 12 columns per level, achieving thorough coverage across different bridge dimensions as shown in Figure 5.

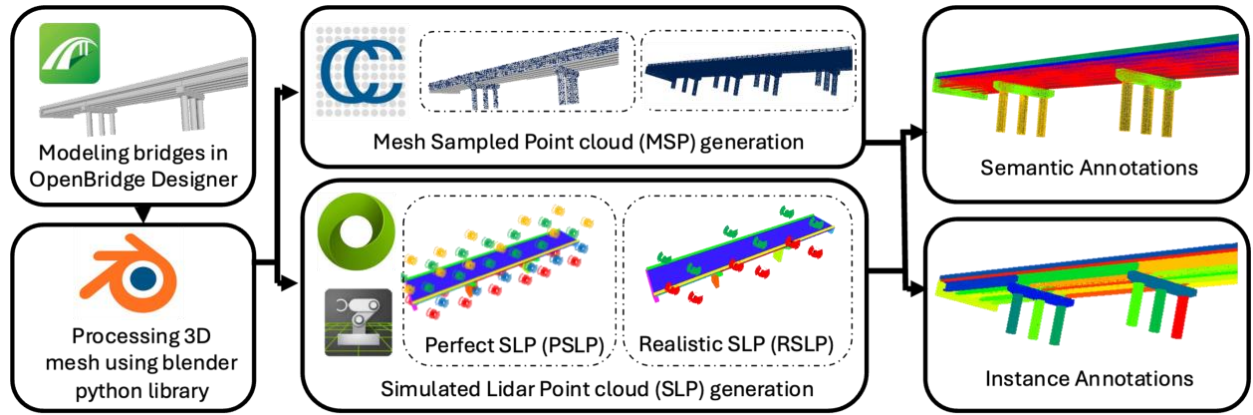

Figure 5. Illustrates three different synthetic bridge point clouds generation processes: Mesh Sampled Point Cloud (MSP), Realistic Simulated LiDAR Point Cloud (RSLP), and Perfectly Simulated LiDAR Point Cloud (PSLP).

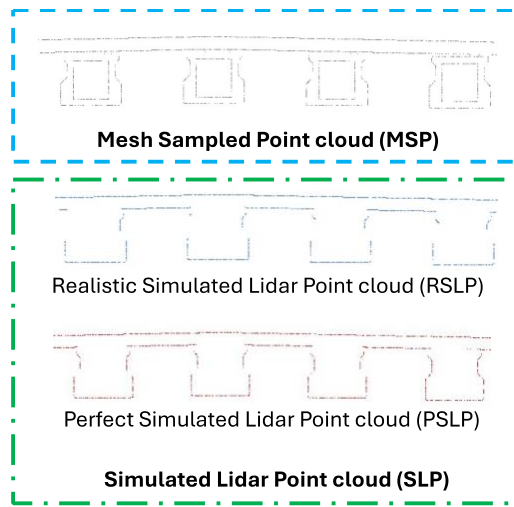

Figure 6. Cross-sections of synthetic point clouds illustrating Mesh Sampled (MSP) with interior detail, Perfectly Simulated LiDAR (PSLP) capturing detailed point cloud, and Realistically Simulated LiDAR (RSLP) mirroring real-world LiDAR capture.

## 2.2. Data pre-processing strategies

The generated synthetic point cloud data undergoes pre-processing steps that include introducing synthetic occlusion and voxelization of the bridge point cloud.

Occlusion is essential for enhancing LiDAR simulation datasets, mimicking real-world scanning challenges like obstructions from bridge components or vegetation. This is achieved by introducing geometric shapes such as cubes, spheres, and prisms inside the bridge point cloud, positioned randomly and sized variably to simulate occlusions. Instead of completely removing occluded points inside those geometric shapes, they are made sparser using a predetermined sparsity factor, reflecting realistic LiDAR data capture from multiple angles. This method ensures a comprehensive dataset for more accurate simulation and analysis, as illustrated in the accompanying Figure 7.

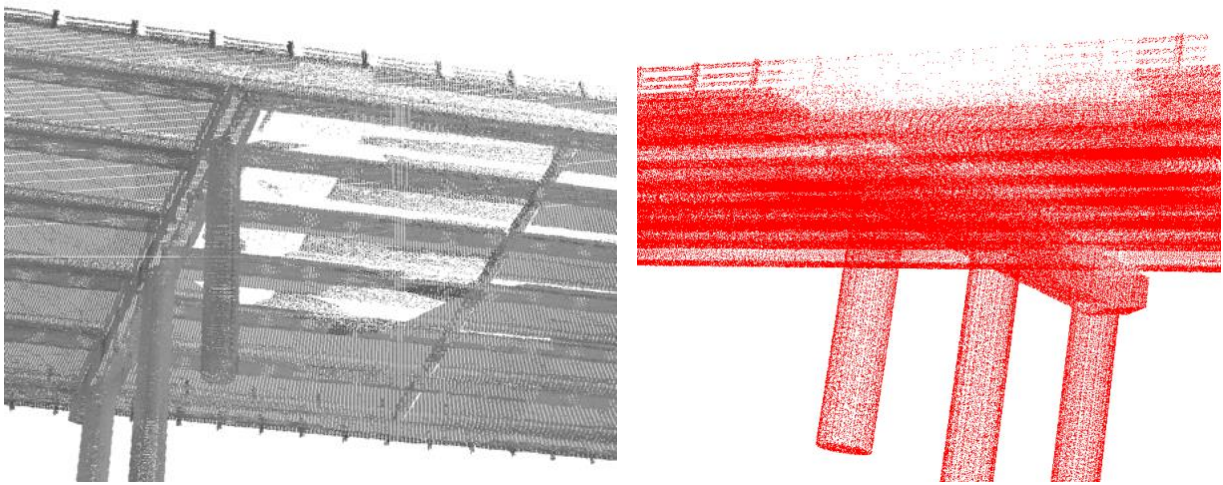

Figure 7. Occlusion representation in real-world bridge point cloud (gray) and its synthetic counterpart (red), with occlusions in synthetic data replicated to closely match real-world scanning conditions.

The real bridge point cloud exhibits non-uniform density, introducing unnecessary patterns, and increasing computational costs for training the segmentation model. To address this issue, voxelization is employed, which represents objects within a regular grid of voxels, effectively normalizing the data density [50]. By adopting a voxel size of 2 cm, we were able to down-sample the point clouds, achieving a uniform density across the datasets.

## 2.3. Data augmentation strategies

To enhance model generalization and performance on real-world data, data augmentation techniques were employed, including a novel sparsity-based occlusion and standard augmentations. We employ standard augmentation techniques such as random scaling, random rotation, cropping, and horizontal flipping to address discrepancies between synthetic and real-world data. Our augmentation pipeline includes scaling within ±10% and rotation around the three principal axes, with limits set to full 360 degrees for the z-axis and smaller ranges for the x and y axes to mimic realistic tilting. The bridge point clouds are cropped into cuboids, further diversifying the data by providing variations in scale and altering spatial contexts.

### 2.4. Instance Segmentation Model

we employed the Mask3D instance segmentation model proposed by Jonas Schult et al. [43], currently recognized as the leading architecture across multiple benchmark datasets, including STPLS3D [51], ScanNet(v2) [52], and S3DIS. The preprocessing involved cropping the point cloud into equal-sized parallelepiped blocks to manage data input size. The colored point cloud data, initially $P \times R_n \times 6$, was then down sampled into voxels ($V \times R_m \times 6$).

The core of Mask3D features a Sparse Feature Backbone using the MinkowskiEngine-based symmetric U-Net architecture for efficient feature extraction [53]. The model also incorporates a transformer decoder with a mask module (MM) that, for each of several queries, predicts a binary mask per instance, refined through a series of layers and mapped to feature layers from the backbone using cross-attention as illustrated in Figure 8. The masks are generated by combining instance features with point feature maps through a dot product, yielding a similarity score converted to a binary mask via sigmoid activation, effectively classifying each instance into one of the classes.

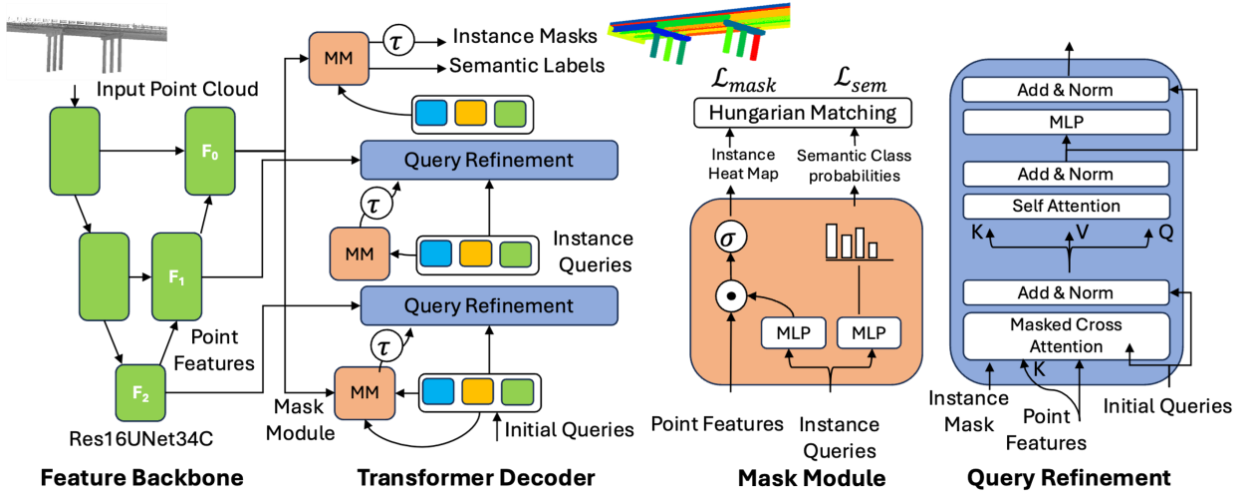

Figure 8. Mask3D architecture of point cloud segmentation network with a Res16UNet34C backbone, Transformer decoder for instance and semantic segmentation, and query refinement stages for enhanced feature extraction and classification Jonas Schult et al. [43].

To establish the correspondence between the predicted and ground truth instances, bipartite graph matching is employed. The cost matrix $C$ is constructed as given by equation (1).

$$C(k, \hat{k}) = \lambda_{dice}\mathcal{L}_{dice}(k, \hat{k}) + \lambda_{BCE}\mathcal{L}_{BCEmask}(k, \hat{k}) + \lambda_{cl}\mathcal{L}_{CE}^{l}(k, \hat{k}) \quad (1)$$

Here, $\mathcal{L}_{dice}$ is the Dice loss, $\mathcal{L}_{BCEmask}$ is the binary cross-entropy loss over the foreground and background of the mask, and $\mathcal{L}_{CE}^{l}$ is the multi-class cross-entropy loss for classification. The weights are set to $\lambda_{dice} = \lambda_{cl} = 2.0$ and $\lambda_{BCE} = 5.0$.

After establishing the correspondence using the Hungarian matching the model optimizes the predicted mask with the following equation (2).

$$\mathcal{L}_{mask} = \lambda_{BCE}\mathcal{L}_{BCE} + \lambda_{dice}\mathcal{L}_{dice} \quad (2)$$

The overall loss for all auxiliary instance predictions is defined in equation (3)

$$\mathcal{L} = \sum_{l}^{L} \mathcal{L}_{mask}^{l} + \lambda_{cl}\mathcal{L}_{CE}^{l} \quad (3)$$

## 2.5. Field Data Collection

The deep instance segmentation model was evaluated with real bridge point cloud data from the Terrestrial Laser Scanner (TLS) and the Photogrammetry. The real point cloud data was collected from a service road bridge over Brays Bayou in Houston, Texas located at coordinates 29°42'44.4" N, 95°22'39.8" W.

### 2.5.1. Point cloud acquisition using Terrestrial Laser Scanner (TLS)

For the real bridge point cloud data acquisition using Terrestrial Laser Scanner (TLS) we used the RIEGL VZ-400 laser scanning system. The scanner settings included a horizontal field-of-view of 360° and a vertical field-of-view of 100° (ranging from -40° to 60°), with an angular scanning resolution of 0.015°. It operated at a scanning frequency of 1010 kHz, capable of measuring distances up to 580 meters with a measurement precision of 2 mm at 100 meters. For accurate georeferencing, GNSS technology, supplemented by three reflectors, facilitated the automated alignment and registration of point clouds from various scanner positions. The scans were conducted from six terrestrial laser scanning (TLS) stations beneath the bridge and four above to capture the entire structure comprehensively. The final aligned and georeferenced point cloud was recorded in the Universal Transverse Mercator (UTM) global coordinate system, totaling 2,475,529 points.

### 2.5.2. Point cloud acquisition using photogrammetry

For the photogrammetric point cloud, the images data was captured by a Skydio 2+ UAV. This process involved taking 7068 images of the bridge, maintaining a 76% side lap and overlap, and a consistent 5ft distance from the surface. Various scanning techniques were utilized: the upper part of the bridge was documented with a 2D downward scan, the lower deck with a 2D upward scan at gimbal angles of 80 and 60 degrees for complete girder visibility, and the piers and pier caps with 3D capture, while keyframe manual mode was used for the front and back. The collected data were then reconstructed into a detailed 3D model using the WebODM software.

The annotation of the bridge point cloud, captured with Terrestrial Laser Scanning (TLS), was carefully performed using the CloudCompare. Each structural component within the point cloud was labeled with corresponding semantic and instance classes. We defined five semantic classes encompassing slabs, barriers, pier caps, piers, and girders. Additionally, unique instance labels were assigned to every individual component for instance annotation as shown in the Figure 9. These thoroughly detailed annotations served as the ground truth for evaluating our model, providing a reliable benchmark for assessing segmentation accuracy.

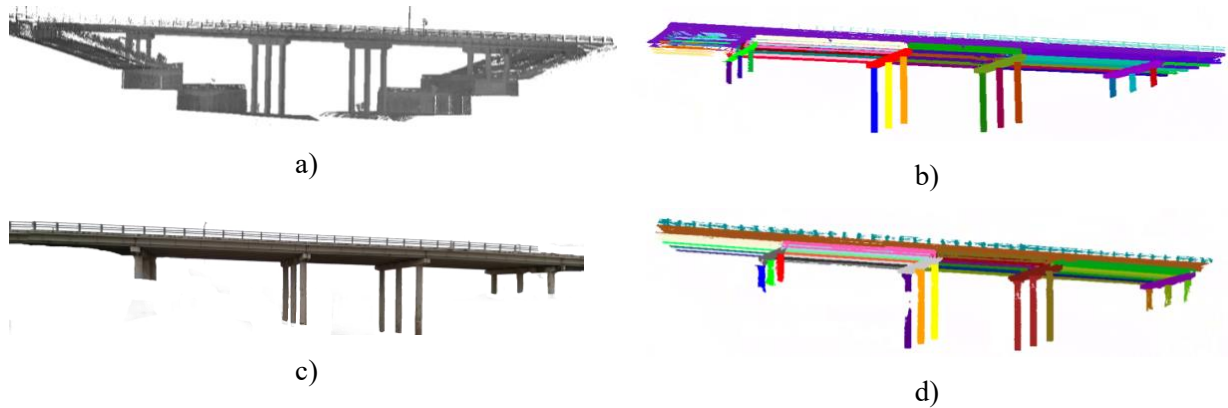

Figure 9. shows the real bridge point clouds and ground truth instance labels: (a) LiDAR-acquired point cloud, (b) instance labels for LiDAR data, (c) photogrammetry-acquired point cloud, and (d) instance labels for photogrammetry data, with distinct colors for structural components.

## 3. Training and Evaluation

The model underwent training using synthetic data and was subsequently evaluated on real data to assess its performance. For hyperparameter tunning, out of the total 60 synthetic bridges generated, 52 were allocated for the training set while the remaining 8 were designated for validation. These validation bridges were carefully selected to represent the entire spectrum of the data distribution. To maintain consistency in our results, the same set of validation bridges was used across all hyperparameter tunning experiments. This approach allowed for consistent comparative analysis of the model's performance under various conditions.

In subsequent experiments to evaluate the "effect of synthetic dataset type on segmentation" and the "effect of occlusion," the point clouds generated from the same set of 52 synthetic bridges (using different techniques) were consistently used for training. For validation, the real LiDAR and photogrammetry point clouds were split into two halves; one half was used for validation, and the other half was reserved for testing.

The training process involved several key steps and configurations to optimize performance. Initially, the dataset underwent preprocessing to prepare it for effective learning. Training was conducted for 70 epochs with an initial learning rate of $10^{-4}$. The total number of epochs was determined based on the stabilization of the mean loss curve, as shown in Figure 10. Extending the training duration beyond this point did not yield better results. We employed a voxel size of 0.2 meters and utilized an NVIDIA GeForce RTX 3090, which resulted in training times ranging from approximately 24 to 36 hours. The AdamW optimizer [54], along with a one-cycle learning rate schedule [55], was implemented to enhance optimization. To foster model robustness and generalization, we incorporated random occlusion in both training and validation datasets, in addition to our standard augmentation techniques.

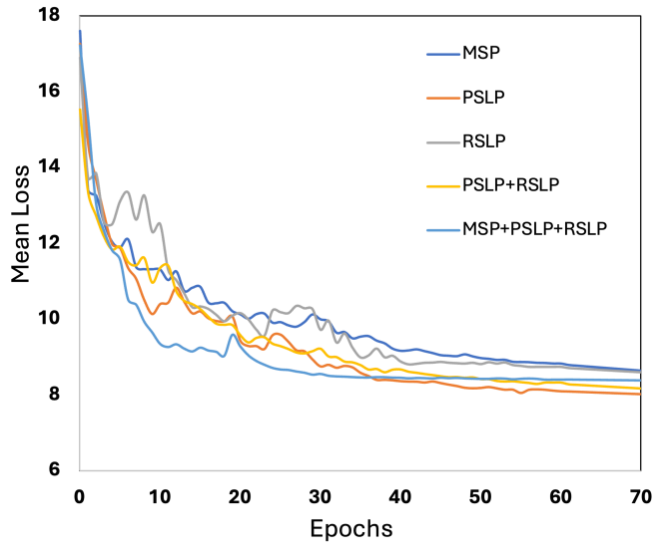

Figure 10. Illustrates the training mean loss on the y-axis and epochs on the x-axis, illustrating a sharp initial decrease in loss followed by gradual stabilization.

The performance of the model trained on synthetic data was evaluated on real LiDAR and photogrammetry point cloud using Average Precision (AP) metrics, specifically mAP, mAP_25%, and mAP_50%. For this evaluation, the model weights corresponding to the peak validation mAP were used to ensure the best possible performance assessment.

$$Precision = \frac{tp}{tp + fp}$$

$$tp = true\ positive$$

$$fp = false\ positive$$

AP: Average precision (AP) calculates the mean precision across all the IoU thresholds except 0.25. this average includes IoU thresholds like 0.5, 0.55, 0.6, etc.

AP25: Calculates the mean AP specifically at 25% IoU threshold.

AP50: Calculates the mean AP specifically at 50% IoU threshold.

## 4. Results and Discussion

This section presents the results of hyperparameter tuning and examines the impact of different synthetic dataset types on the instance segmentation of bridge point clouds. Unlike synthetic bridge point clouds, which accurately represent bridge geometry, real bridges exhibit occlusions that can affect segmentation performance. To evaluate the impact of occlusion, we conducted experiments introducing occlusion to all synthetic data types. These experiments aimed to assess how occlusion influences the model's segmentation performance.

### 4.1. Hyperparameter tuning

Hyperparameter tuning played an important role in optimizing our model's performance. We carefully selected a validation set from our synthetic dataset, comprising a fixed size of 8 bridges out of 60 PSLP synthetic dataset to ensure coverage across the entire data distribution. No cross-validation was used; instead, we consistently applied the same validation set to evaluate various hyperparameter adjustments. Key hyperparameters that underwent tuning included voxel size, the epsilon parameter for DBSCAN clustering (dbscan eps), the minimum number of points required to form a cluster in DBSCAN (dbscan min no of points), and variations in the coloration of the point cloud, ranging from uniform white to random and varying colors along the height. Notably, during the tuning phase, occlusion was deliberately not applied to isolate and better understand the effects of other hyperparameter adjustments on model performance.

**Voxel Size:**

Understanding the impact of voxel size on model performance is critical due to its influence on resolution and computational demands [56]. In our study, point cloud data was down sampled into voxels, where smaller sizes increase resolution but also computational costs. We tested voxel sizes from 0.1m to 0.3m and found that while a 0.1m size did not improve model precision and caused memory issues, a 0.3m size led to significant loss in semantic detail. As illustrated in Table 1 the optimal balance was achieved with a 0.2m voxel size, providing the highest mean AP at 74.2 %, efficiently balancing computational efficiency with sufficient resolution to capture bridge structures accurately. This makes a 0.2m voxel size ideal for our application, as demonstrated by the AP variations across sizes.

Table 1. Impact of voxel size on instance segmentation precision.

| *Voxel Size* | *mAP* | *mAP$_{50}$* | *mAP$_{25}$* |
|:---:|:---:|:---:|:---:|
| 0.1 | 0.713 | 0.768 | 0.782 |
| **0.2** | **0.74** | **0.802** | **0.811** |
| 0.3 | 0.574 | 0.748 | 0.759 |

**DBSCAN parameters:**

Studying the sensitivity of DBSCAN parameters—epsilon ($\varepsilon$) and minimum number of points (MinPts)—is essential for optimizing the Mask3D model's instance segmentation performance. This model utilizes DBSCAN to refine the segmentation by splitting merged instances into distinct masks. Our sensitivity analysis involved adjusting $\varepsilon$ values from 0.5 to 10 and MinPts from 1 to 4, revealing optimal performance at an epsilon ($\varepsilon$) of 0.92 and MinPts of 4. These settings yielded the highest precision metrics (mAP 0.742, AP$_{50}$ 0.8, AP$_{25}$ 0.811) as shown in the Table 2 and Table 3, indicating that a higher MinPts threshold helps form more defined and conservative clusters, thus improving accuracy. During experiments for epsilon ($\varepsilon$), the default MinPts value of 4 was used. Once the optimal epsilon was determined, $\varepsilon = 0.92$ was used for subsequent experiments to find the optimal MinPts.

Table 2. Impact of varying DBSCAN epsilon values on model performance across three precision metrics.

| *DBSCAN eps ($\varepsilon$)* | *mAP* | *mAP$_{50}$* | *mAP$_{25}$* |
|:---:|:---:|:---:|:---:|
| 0.5 | 0.711 | 0.76 | 0.781 |
| **0.92** | **0.742** | **0.802** | **0.811** |
| 2 | 0.738 | 0.78 | 0.8 |
| 4 | 0.724 | 0.771 | 0.79 |
| 10 | 0.736 | 0.78 | 0.802 |

Table 3. Influence of minimum number of points (MinPts) in DBSCAN clustering on the segmentation.

| *DBSCAN MinPts* | *mAP* | *mAP$_{50}$* | *mAP$_{25}$* |
|---|---|---|---|
| 1 | 0.64 | 0.732 | 0.778 |
| 2 | 0.729 | 0.751 | 0.797 |
| 3 | 0.731 | 0.786 | 0.795 |
| **4** | **0.742** | **0.802** | **0.811** |

**Color of the point cloud:**

We evaluated the effect of different color schemes on the performance of the Mask3D model, which segments point clouds based on color and geometric features. The experiment compared three schemes: uniform white, random RGB, and varying colors along the height (Z-axis). Initially, point clouds had no color, making this exploration crucial for understanding how color impacts segmentation. Uniform white was used as the baseline, where performance was standard. Introducing random RGB colors slightly reduced the model's accuracy (mAP), likely due to the challenges in segment differentiation, results presented in Table 4. Conversely, applying varying colors along the height slightly improved mAP, providing additional contextual clues that enhanced segmentation.

Table 4. Impact of color variation on the performance of the model, with varying colors along the height showing slightly favorable results.

| *Point cloud color* | *mAP* | *mAP$_{50}$* | *mAP$_{25}$* |
|---|---|---|---|
| White | 0.732 | 0.798 | 0.811 |
| Random RGB | 0.71 | 0.782 | 0.801 |
| Varying color along the height | 0.741 | 0.807 | 0.815 |

These findings shown in the Table 4, underscore the importance of color in training synthetic point cloud datasets, with strategic coloring along the Z-axis proving beneficial for improved segmentation accuracy.

The models inference on validation data after optimizing the hyperparameters of the 3D point cloud instance segmentation model are presented in the Figure 11.

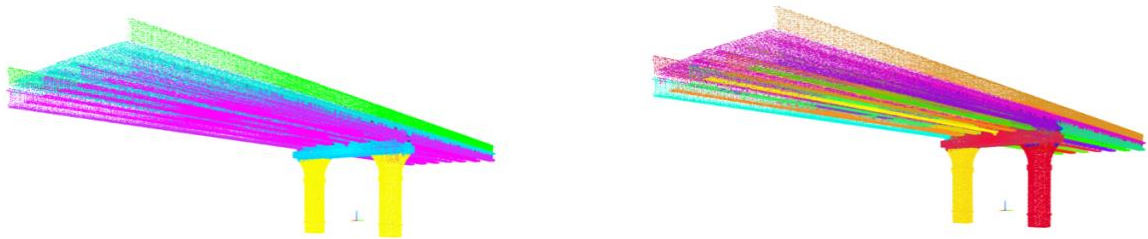

Figure 11. Results of semantic (left) and instance (right) segmentation on a synthetic bridge point cloud from validation data, following hyper-parameter tuning.

### 4.2. Effect of synthetic dataset type on segmentation

The instance segmentation performance of the deep learning model trained on various synthetic datasets and tested on real-world bridge LiDAR point cloud shows significant variance depending on the type of synthetic data used, as detailed in Table 5. As outlined in the training and evaluation section, half of the real LiDAR point cloud data was used for validation and the other half for testing. When trained solely on Mesh Sampled Point Cloud (MSP), the model displayed notably low performance when tested on LiDAR point cloud (mAP of 0.002, $mAP_{50}$ of 0.007, and $mAP_{25}$ of 0.43). The poor results can be attributed to MSP's significant deviations from real-world data, including differences in the shape of structural components and the inclusion of points sampled from the interiors of structural members—features not typically captured by real-world LiDAR.

In contrast, training with Perfect Simulated LiDAR Point Cloud (PSLP) and Realistic Simulated LiDAR Point Cloud (RSLP) individually showed considerable improvements. PSLP, which simulates a dense grid of LiDAR sensors to avoid occlusions and captures nearly every part of the bridge, achieved a mAP of 0.790. RSLP, being the closest to real-world point clouds, further improved model performance, achieving a mAP of 0.89. The combination of PSLP and RSLP yielded the highest performance (mAP of 0.964, $mAP_{50}$ of 0.992, $mAP_{25}$ of 0.992) by providing comprehensive learning on both exact component geometry and realistic LiDAR sensor placement.

Table 5. Comparative performance of deep learning model trained on different synthetic datasets and tested on real-world bridge point clouds.

| *Exp #* | *Training data type* | *Test data* | *mAP* | *$mAP_{50}$* | *$mAP_{25}$* |
|---|---|---|---|---|---|
| 1 | MSP | LiDAR PC | 0.002 | 0.007 | 0.43 |
| 2 | PSLP | LiDAR PC | 0.790 | 0.983 | 0.983 |
| 3 | RSLP | LiDAR PC | 0.890 | 0.982 | **0.993** |
| 4- | PSLP+RSLP | LiDAR PC | **0.964** | **0.992** | 0.992 |

After evaluating the model's performance with the real LiDAR bridge point cloud, we tested the same models with a real photogrammetry bridge point cloud. The average precision values were lower compared to those for the LiDAR point cloud, as shown in Table 6. This discrepancy is attributed to the fundamentally different nature of the photogrammetry point cloud, which is less accurate and exhibits more irregularities than the LiDAR point cloud [57]. These irregularities make inference challenging for the model, which was trained on synthetic bridge point clouds that are more precise and better represent the bridge geometry. A similar trend was observed for the photogrammetry bridge point cloud, with the best performance achieved using the PSLP+RSLP dataset (mAP of 0.638, $mAP_{50}$ of 0.903, $mAP_{25}$ of 0.903).

Table 6. Comparative performance of deep learning model trained on different synthetic datasets and tested with photogrammetric bridge point clouds.

| *Exp #* | *Trained with* | *Test* | *mAP* | *$mAP_{50}$* | *$mAP_{25}$* |
|---|---|---|---|---|---|
| 1 | MSP | Photogrammetry | 0.001 | 0.005 | 0.10 |
| 2 | PSLP | Photogrammetry | 0.431 | 0.615 | 0.746 |
| 3 | RSLP | Photogrammetry | 0.595 | 0.890 | **0.904** |
| 4 | PSLP+RSLP | Photogrammetry | **0.638** | **0.903** | 0.903 |

## 4.3. Effect of Occlusion

Introducing occlusion to the synthetic dataset for training the instance segmentation model is crucial because real bridge point clouds significantly differ from synthetic ones. Real bridge point clouds exhibit occlusions due to various factors such as vegetation, areas outside the vertical field of view, cars, objects, humans, and girders obstructing the view of the Terrestrial Laser Scanner (TLS). To simulate these conditions, occlusions are introduced to the synthetic bridge point clouds as described in section 3.2. For these experiments, the training dataset was doubled compared to previous experiments. For instance, in the PSLP+Occ_N% scenarios, the training data included 52 original bridges without occlusion and an additional 52 bridges with occlusion, resulting in a total of 104 synthetic bridge point clouds.

The model is first trained on the PSLP datasets with occlusion at various sparsity factors and tested against the LiDAR point cloud, as shown in the Table 7. The results indicate that optimal performance is not achieved by removing all points within the occlusion geometry, as different parts of the object may be visible from various stations. Instead, the optimum occlusion sparsity rate should be identified, or the training data should be processed with various sparsity factors. It was observed that for the PSLP a 60% sparsity factor yielded the highest performance (mAP 0.920, $mAP_{50}$ 0.989, and $mAP_{25}$ 0.991) followed by decline at 80% sparsity, indicating the importance of optimum sparsity rate of occlusion in training data.

Table 7. Comparative performance of deep learning model trained with PSLP data with various occlusion sparsity factors and tested with the real bridge point clouds from LiDAR.

| *Sparsity factor* | *Test* | *mAP* | *$mAP_{50}$* | *$mAP_{25}$* |
|---|---|---|---|---|
| 20% | LiDAR PC | 0.804 | 0.956 | 0.979 |
| 40% | LiDAR PC | 0.664 | 0.889 | 0.924 |
| 60% | LiDAR PC | **0.920** | **0.989** | **0.991** |

| | 80% | LiDAR PC | 0.763 | 0.822 | 0.955 |
|---|---|---|---|---|---|

After optimizing the sparsity factor, the optimum rate of 60% was used for all the preceding training datasets, and the model's performance was evaluated using real bridge LiDAR (Table 8) and photogrammetry (Table 9) point clouds. The results, as shown in Table 8, indicate that the PSLP+RSLP+occ_60% dataset yielded the highest (mAP 0.759, $mAP_{50}$ 0.929, and $mAP_{25}$ 0.965) values for the LiDAR point cloud, due to the introduction of occlusion.

Table 8. Comparative performance of deep learning model trained on different synthetic datasets with 60% occlusion sparsity rate and tested on bridge point clouds from LiDAR.

| Exp # | Trained with | Test | mAP | $mAP_{50}$ | $mAP_{25}$ |
|---|---|---|---|---|---|
| 1 | MSP+occ 60% | LiDAR PC | 0.008 | 0.012 | 0.17 |
| 2 | PSLP+occ 60% | LiDAR PC | 0.920 | 0.989 | 0.991 |
| 3 | RSLP+occ 60% | LiDAR PC | 0.911 | 0.984 | 0.997 |
| 4 | PSLP+RSLP+occ 60% | LiDAR PC | **0.978** | **0.999** | **0.999** |

In contrast, the model's performance declined when occlusion was added and evaluated with the photogrammetry point cloud. This decrease can be attributed to the fundamentally different nature of photogrammetry point clouds. Unlike LiDAR point clouds, photogrammetry point clouds have very little occlusion because the image data was collected extensively, allowing for a complete reconstruction of the bridge without missing any occluded areas. Therefore, adding occlusion to the training data did not positively impact the model's inference for photogrammetry point cloud.

Table 9. Performance of deep learning model trained on different synthetic datasets with 60% occlusion sparsity rate and tested on bridge point clouds from photogrammetry.

| Exp # | Trained with | Test data | mAP | $mAP_{50}$ | $mAP_{25}$ |
|---|---|---|---|---|---|
| 1 | MSP+occ 60% | Photogrammetry | 0.004 | 0.07 | 0.13 |
| 2 | PSLP+occ 60% | Photogrammetry | **0.496** | **0.826** | 0.843 |
| 3 | RSLP+occ 60% | Photogrammetry | 0.458 | 0.603 | 0.697 |
| 4 | PSLP+RSLP+occ 60% | Photogrammetry | 0.475 | 0.763 | **0.928** |

Based on the above experiments, this study proposes two important considerations to maximize the model performance when using synthetic point cloud data for instance segmentation of bridge point clouds, including i) using the combination of PSLP and RSLP datasets for photogrammetry point clouds, and ii) addition of optimal occlusion to the training data improves the model performance significantly in case of LiDAR point cloud. The inference of LiDAR and photogrammetry point clouds with the proposed techniques are illustrated in Figure 12.

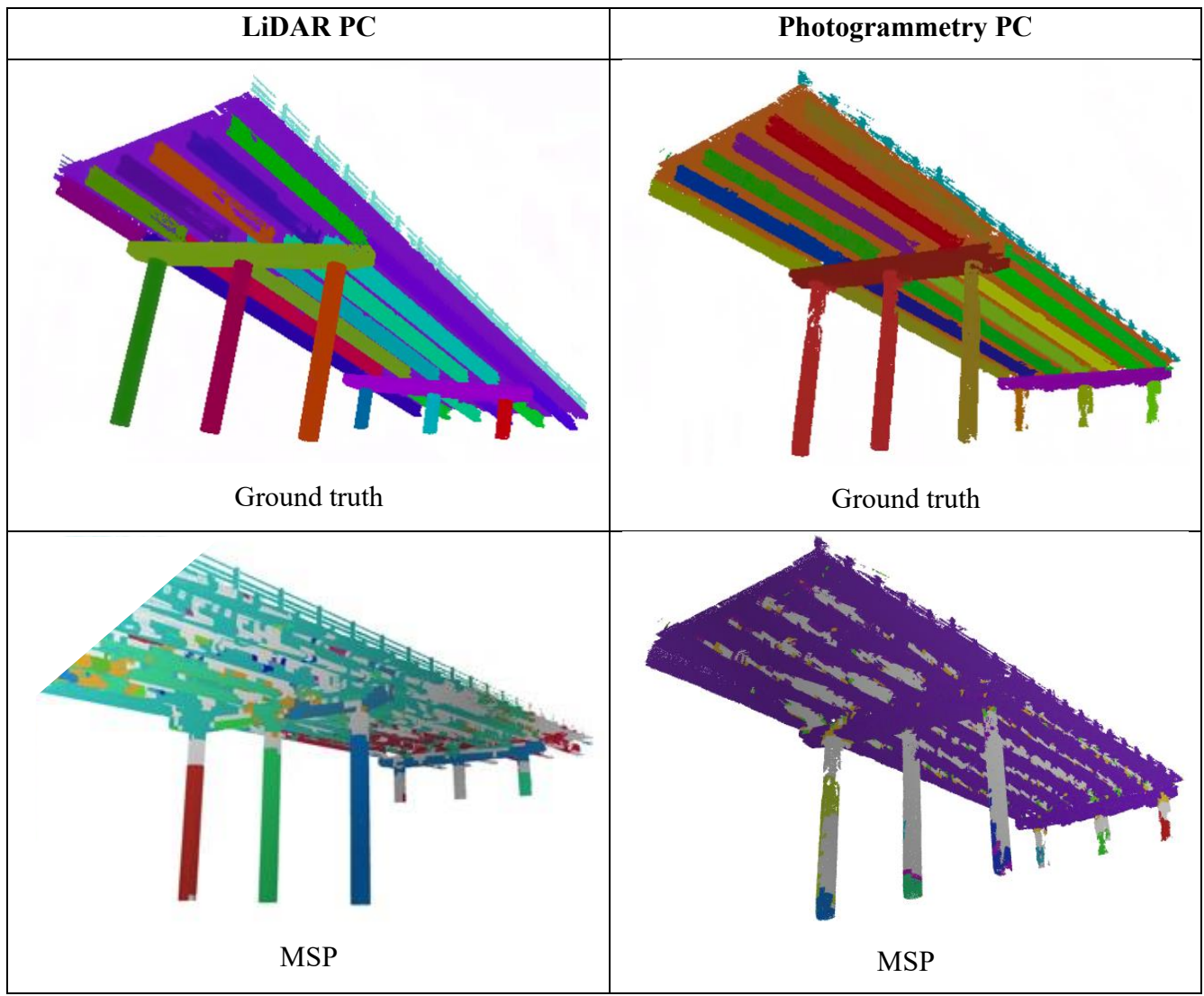

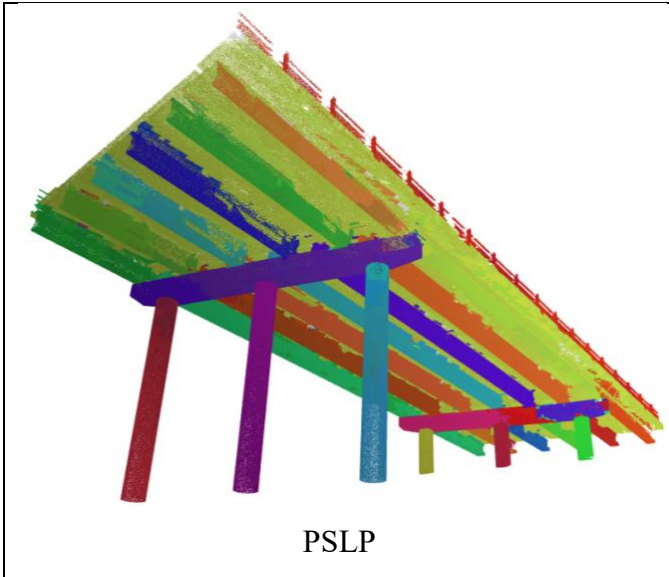
PSLP

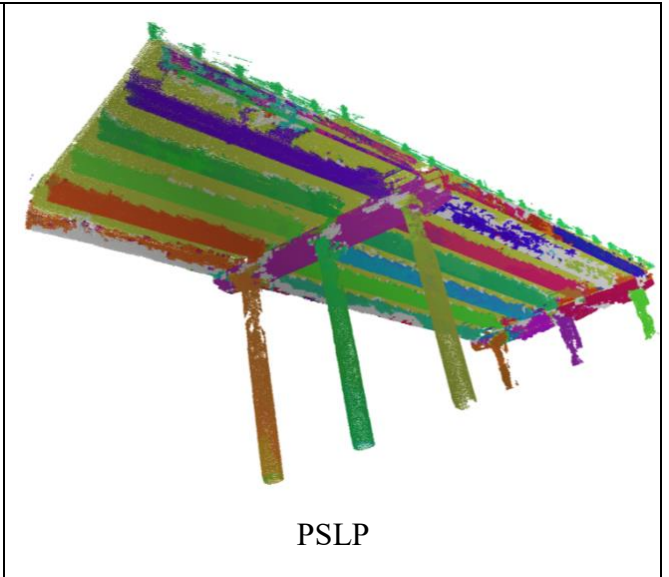
PSLP

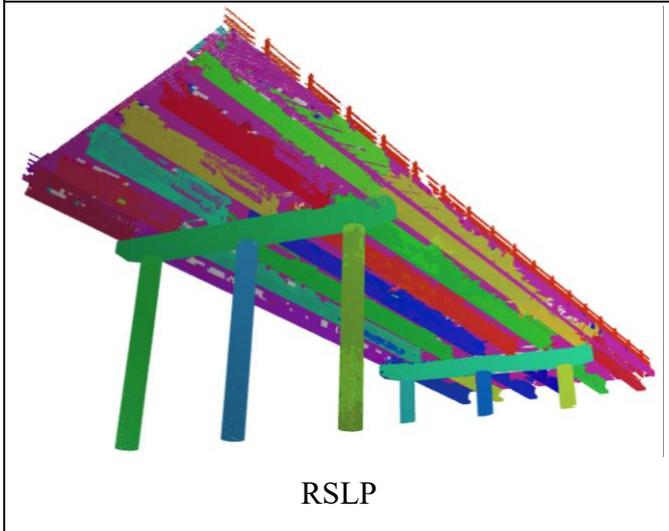
RSLP

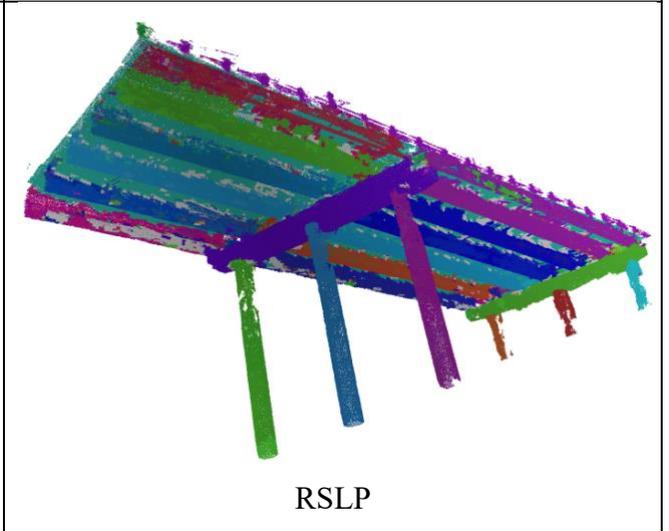
RSLP

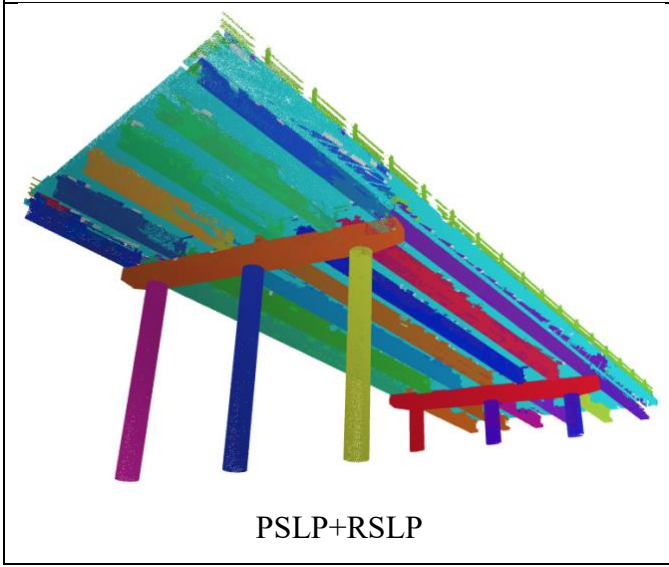
PSLP+RSLP

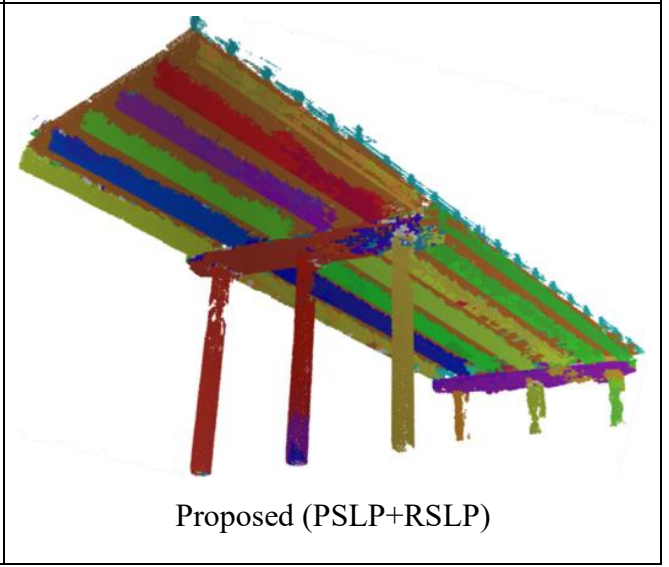
Proposed (PSLP+RSLP)

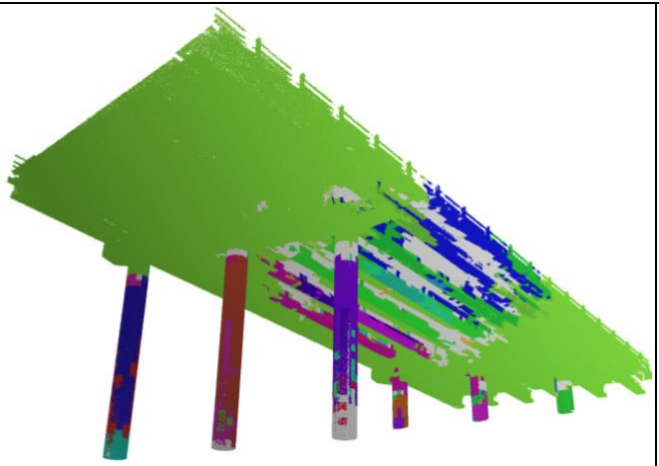
MSP+Occ60

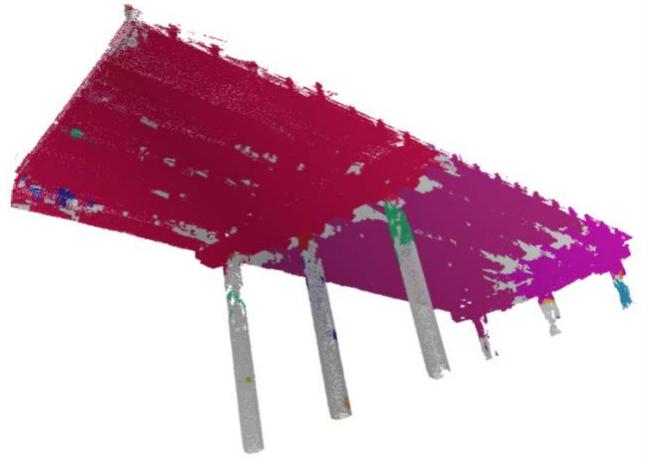
MSP+Occ60

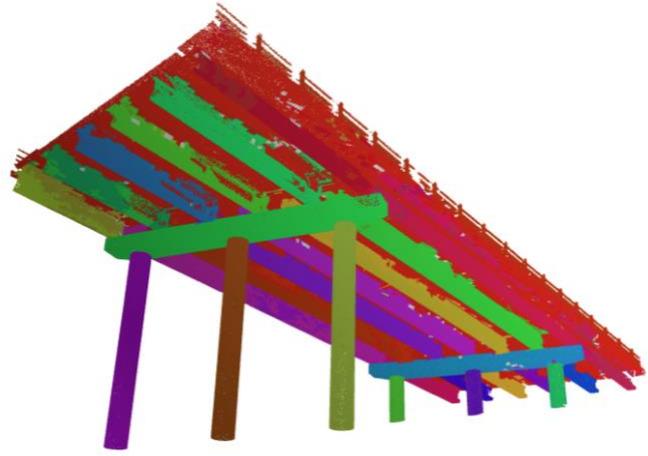
PSLP+Occ60

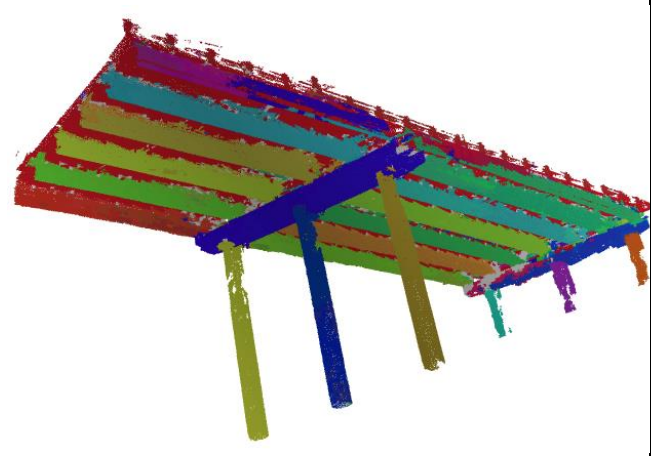
PSLP+Occ60

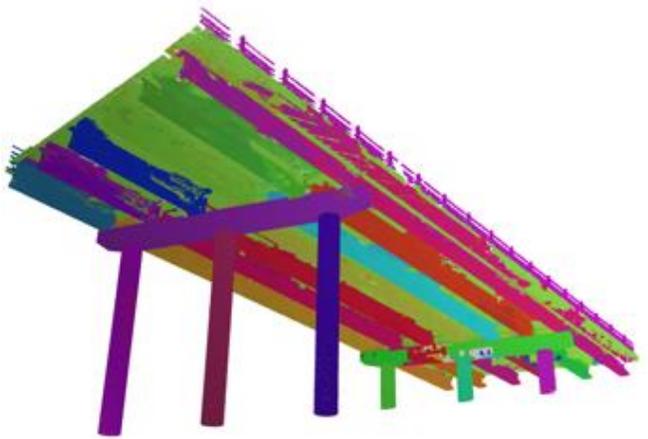
RSLP+Occ60

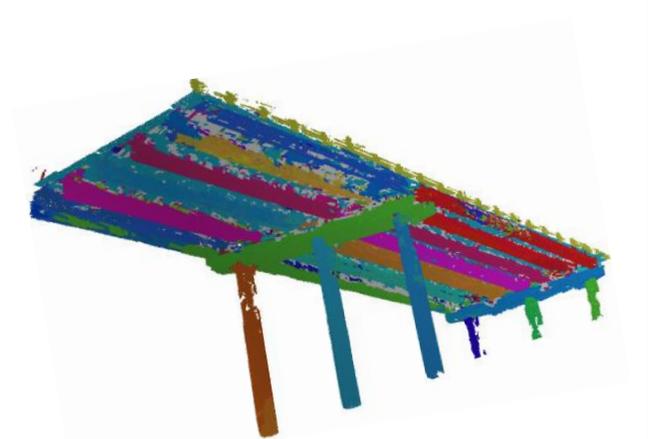
RSLP+Occ60

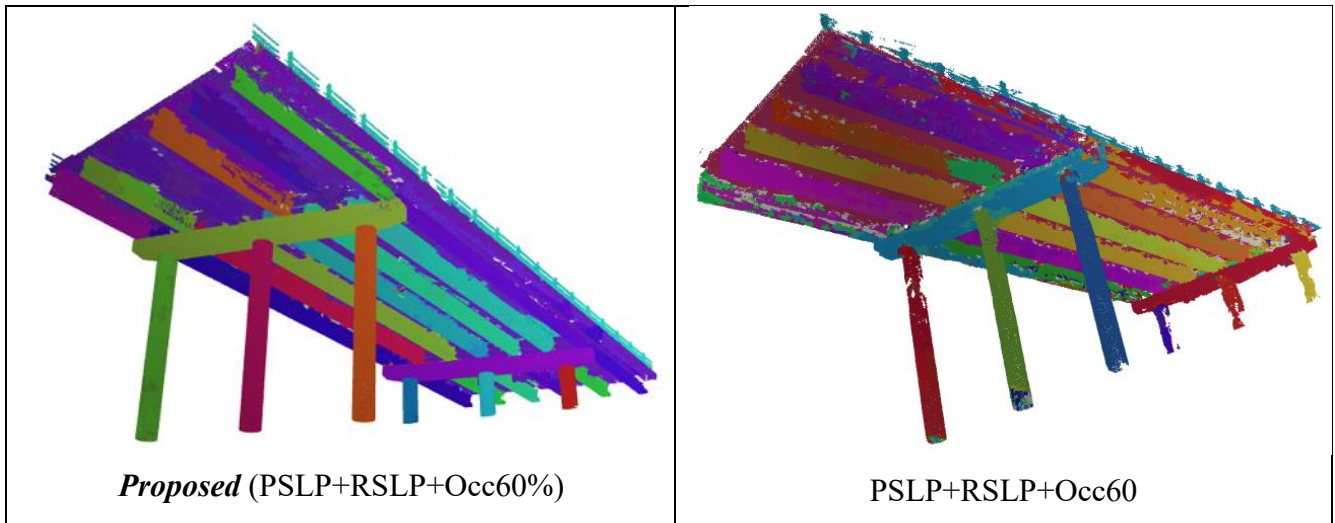

| *Proposed* (PSLP+RSLP+Occ60%) | PSLP+RSLP+Occ60 |

Figure 12. Proposed instance segmentation techniques for LiDAR and photogrammetry real bridge point clouds.

## 5. Conclusion

This research proposes a novel methodology for deep learning-based instance segmentation of structural components in real RC bridge point clouds by using models trained on synthetically generated RC bridge point clouds. To demonstrate our proposed methodology, we developed and evaluated three datasets each with the same 60 bridges but distinct approaches for sampling points from 3D bridge models for datasets, namely Mesh Sampled Point Clouds (MSP), Perfect Simulated LiDAR Point Clouds (PSLP), and Realistic Simulated LiDAR Point Clouds (RSLP). The latter two were developed by densely (PSLP) or practically (RSLP) placing virtual LiDAR sensors around bridge models, respectively, with RSLP closely mirroring real-world accessible locations.

Our findings indicate that synthetic data is highly suitable for the training of point-cloud instance segmentation models. Specifically, training the proposed deep network performed best on real LiDAR point cloud when combining both the PSLP and RSLP data and pre-processed with optimum occlusion (60% sparsity), achieving a mAP of 0.978, $mAP_{50}$ of 0.999 and $mAP_{25}$ of 0.999. This performance is be attributed to the fact that the RSLP provides a close representation of the real data, the PSLP also provides additional information on how a perfect point cloud might look, and 60% sparsity occlusions resemble real-world occlusion patterns observed in our dataset.

The combination of these three processes resulted in improved generalizability to real point clouds compared to applying any process in isolation. For the real photogrammetry point cloud, the highest performance (mAP 0.638, mAP_50 0.903, and mAP_25 0.903) was achieved when the model was trained with the combination of PSLP and RSLP data without occlusion pre-processing. likely due to the minimal occlusion in photogrammetry data. As expected, MSP data proved inadequate for training models due to its discrepancy from real-world scenarios. Our findings indicate that varying or randomizing point cloud colors does not significantly impact the model's performance, and that finer voxel resolutions do not necessarily equate to better results. This study presents a framework for synthetic bridge point cloud dataset creation and utilizing it for instance segmentation of bridge components. The implementation of this framework will be valuable in automating the bridge inspection process by aiding in bridge element rating and to create geometric digital twin of bridges.

## 6. Funding

This research is supported by the Texas Department of Transportation (TxDOT) under project number TxDOT 0-7181.